\pdfoutput=1

\documentclass[11pt]{article}
\usepackage{authblk}
\usepackage{acl}

\usepackage{times}
\usepackage{latexsym}
\usepackage{amsfonts}
\usepackage{comment}
\usepackage[T1]{fontenc}

\usepackage[utf8]{inputenc}

\usepackage{microtype}

\usepackage{graphicx}
\usepackage{booktabs}
\usepackage{float}

\title{Dynamic Embedded Topic Models: properties and recommendations based on diverse corpora}

\author[1,2]{Elisabeth Fittschen}
\author[1]{Bella Xia}
\author[1]{Leib Celnik}
\author[3]{Paul Dilley}
\author[1]{Tom Lippincott}
\affil[1]{Johns Hopkins University}
\affil[2]{University of Hamburg}
\affil[3]{University of Iowa}

\begin{document}
\maketitle
\begin{abstract}
We measure the effects of several implementation choices for the Dynamic Embedded Topic Model, as applied to five distinct diachronic corpora, with the goal of isolating important decisions for its use and further development. We identify priorities that will maximize utility in applied scholarship, including the practical scalability of vocabulary size to best exploit the strengths of embedded representations, and more flexible modeling of intervals to accommodate the uneven temporal distributions of historical writing. Of similar importance, we find performance is not significantly or consistently affected by several aspects that otherwise limit the model's application or might consume the resources of a grid search.
\end{abstract}
\nocite{*}

\section{Background}
The Dynamic Embedded Topic Model (DETM, \citet{dieng:2019}) has proven useful for a variety of applied research that complements established practices and hypotheses of humanistic scholarship related to historical semantic change \cite{sirin:2024,cassotti:2024}. DETM provides an interpretable, minimally-supervised reflection of semantic change that combines the representational advantages of word embeddings \cite{dieng:2020}\footnote{Released before and cited by the DETM paper, but formally appearing in TACL shortly after.} with explicit modeling of distributional shifts for both topic proportions and words \cite{blei:2006}. However, it is considerably more complex, and its behavior less-explored, than the original Bayesian network \cite{blei:2003} and its variants. Due to its potential utility for the vast amount of humanistic scholarship that hinges on the development of concepts or the waxing and waning of themes in the written record across time, it is important to begin mapping this behavior to eventually establish best practices and architectural improvements. The main contribution of this paper is an initial map, and a handful of research directions that map suggests to be more or less promising.

\section{Materials and Methods}

\subsection{Corpora}

\begin{table}[H]
\centering
\begin{tabular}{lrrr}
\hline
Name & Docs & Tokens & Date range \\ 
\hline
acl & 10628 & 42m & 1965--2006\\
greek & 1696 & 28m & -725--650\\
latin & 574 & 16m & -219--497\\
scifi & 2553 & 92m & 1870--1929\\
un & 7314 & 21m & 1970--2014 \\
\hline
\end{tabular}
\caption{Corpora used in this study, with document count, token count, and date range.}
\label{table:corpora}
\end{table}

We assembled five diachronic text corpora, including those from the original DETM paper (\texttt{acl} and \texttt{un}), summarized in Table \ref{table:corpora}. The \texttt{greek} corpus is the First1KGreek \cite{first1kgreek} materials with dates from the Leuven Databast of Ancient Books \cite{ldab}, the \texttt{latin} corpus is derived from the Perseus Project, and the \texttt{scifi} corpus was assembled from the HathiTrust \cite{htc} to contain contemporaneous science fiction, popular science, and research. We release these standardized datasets with the paper.\footnote{https://huggingface.co/datasets/ANONYMIZED}

\subsection{Sub-documents and embeddings}
Each corpus was randomly divided into training, validation, and test sets at the level of complete documents and in proportions of \( 0.8 \), \( 0.1 \),
and \( 0.1 \), respectively.  The documents were then split into \emph{sub-documents} of at most 100 words each. The Word2Vec Skip-gram embedding models \cite{mikolov:2013} were trained in the combined training and development sets.

\subsection{Model hyper-parameters}
The model choices tested in this paper, and the motivation for considering them, are: 

\textbf{Recomputing time window statistics:} in the original implementation, summary word statistics for the mixture-prior inference network were recounted for different stages of use (training, validation, and ultimately, test). This raises a question of interpretation: do these priors in the graphical model correspond to the historical mixture of topics in each window, or something tailored to the specific corpus under consideration at each stage? On a more immediate practical level, at test time one might reasonably want to apply the model to a single document, or author, or a set of materials chosen with by some criterion that makes their temporal-semantic distributions unnatural. Rather than diving immediately into the pros and cons, we test whether there is substantial performance difference between this recomputation, and simply using the training statistics as the model's permanent foundation.

\textbf{Weighting components of the loss:} in the original implementation, the KL-divergence of the document topic-mixtures, and the NLL from reconstruction, are reweighted by the ratio of total training documents to batch size. Using a global statistic to modulate batch loss has drawbacks, such as in the case of continued training or fine-tuning, so we test whether it is an important factor.

\textbf{Deltas for time window evolution:}
The random walks for the changes in mixture-distribution priors and topic embeddings both use a "delta" parameter, in the original paper set to the same value. We experiment with varying the ratios of delta for the mixture walk to the topic walk from $\frac{1}{9}$ to $\frac{9}{1}$.

\textbf{Time window size:}
We track performance for a range of window counts, from the minimum of \( 2 \) to a maximum of \( 32 \), by which point all corpora have some empty windows. This tests the effect of our smoothing approach, but more importantly, sets the stage for our future work on continuous extension of the model.

\textbf{Vocabulary size:}
A major advantage DETM is handling long-tail vocabulary due to data-efficient training of context-independent embeddings. We test maximum vocabulary sizes of \( 10k \), \( 20k \), and \( 80k \).

\textbf{Topic count:}
We hypothesize that, due to the richer representations of rare vocabulary and phenomena afforded by embeddings, DETM may be able to meaningfully populate larger inventories of topics, and test from \( 2 \) to \( 160 \).

\subsection{Experiments}

For each free hyper-parameter, we sweep (or, if discrete and finite, enumerate) its possible values within a tractable range, training a corresponding model for each corpus, while holding all other hyper-parameters to their defaults (see Table \ref{table:defaults}). We report the per-word negative log-likelihood (NLL) on the test set:

\[ \frac{\sum_{d \in D}{\sum_{w \in d}log(\sum_{t \in T}{{P(w|t) \times P(t|d_w)}})}}{total\_words} \]

Previous work \cite{hoyle:2021,chang:2009} has demonstrated the pitfalls of treating NLL as a proxy for human judgment of model quality, a challenging goal that is still debated. We have conveniently decided to focus on basic training and convergence as the first step towards understanding DETM's dynamics, for which NLL provides simple insight. We do however include normalized pointwise mutual information (NPMI) coherence values \cite{bouma:2009} in Appendix \ref{appendix:coherence}, characterize its relationship to NLL, and expand on our reasoning for not focusing on it in this context.
\footnote{Our experimental code is available at \url{https://github.com/ANONYMIZED}}


\subsection{Windows with no observations}
Because empty temporal windows lead to division-by-zero errors for the input to the topic mixture inference network, and such windows inevitably occur unless the choice of window size is heavily restricted, we implement a smoothing policy for that calculation: each empty window is treated as an even mixture of the first non-empty windows to its left and right (or identical to its sole neighbor, if the first or last window). This interpolation is only used to smooth the mixture priors, the reconstruction loss still treats such windows as containing no direct observations.

\subsection{Significance of scores}
While computational constraints prevented doing so for each corpus, we also trained \( 20 \) models on the ACL corpus using the same parameters but different random seeds for data splitting and training. This resulted in a Bessel-corrected estimate of \( 2\sigma = 0.03 \) which may give a sense of significance for our reported numbers. We format in \textbf{bold} the best performance in a table row unless it doesn't meet this threshold.

\section{Results and discussion}
\label{sec:results}

\begin{table}[H]
\centering
\begin{tabular}{lrr}
\toprule
& \multicolumn{2}{c}{Recompute} \\
Corpus & False & True \\
\midrule
acl & 7.05 & 7.04 \\
greek & 6.71 & 6.69 \\
latin & 7.14 & 7.15 \\
scifi & 7.13 & 7.12 \\
un & 7.12 & 7.11 \\
\bottomrule
\end{tabular}
\caption{Average test word NLL with and without recomputing RNN input. None of the corpora show a significant difference.}
\label{table:recompute}
\end{table}

\begin{table}[H]
\centering
\begin{tabular}{lrr}
\toprule
& \multicolumn{2}{c}{Reweight} \\
Corpus & False & True \\
\midrule
acl & \textbf{7.06} & 7.10 \\
greek & \textbf{6.70} & 6.88 \\
latin & \textbf{7.16} & 7.37 \\
scifi & 7.14 & 7.11 \\
un & 7.08 & \textbf{7.01} \\
\bottomrule
\end{tabular}
\caption{Average test word NLL with and without reweighting components of the loss function. Three out of the four corpora showing significant differences favor not reweighting the loss.}
\label{table:reweight}
\end{table}

Recomputing the topic-mixture priors using the particular temporal vocabulary counts for the inference-time documents (i.e. validation and test) showed no significant effect on performance (Table \ref{table:recompute}). Reweighting the loss function based on the ratio of batch to total training size showed moderate performance differences (Table \ref{table:reweight}), but not consistently for the better or worse. For simplicity, it therefore seems best \emph{not} to employ these unless they are further justified, theoretically or empirically. This has the advantage of removing the two needs for tracking global statistics when operating on the batch level.

\begin{table}[H]
\centering
\small
\begin{tabular}{lrrrrrrr}
\toprule
& \multicolumn{7}{c}{Topic Count} \\
Corpus & 2 & 5 & 10 & 20 & 40 & 80 & 160 \\
\midrule
acl & 7.77 & 7.53 & 7.36 & 7.19 & 7.09 & 6.99 & \textbf{6.94} \\
greek & 7.28 & 7.08 & 6.92 & 6.83 & 6.72 & \textbf{6.61} & 6.71 \\
latin & 7.63 & 7.40 & 7.29 & 7.20 & 7.14 & 7.14 & \textbf{7.11} \\
scifi & 7.82 & 7.57 & 7.37 & 7.24 & 7.14 & \textbf{7.13} & 7.16 \\
un & 7.70 & 7.42 & 7.25 & 7.14 & \textbf{7.10} & 7.15 & 7.32 \\
\bottomrule
\end{tabular}
\caption{Average test word NLL across a range of topic counts.}
\label{table:topic_count}
\end{table}

Table \ref{table:topic_count} shows that the model continues to improve with larger topic inventories, in some cases still improving at the top of our tested range. A question for future work is how far this extends, and why some corpora have earlier optima at \( 40 \) or \( 80 \). It may be due to their fundamental topical complexity, but could also be due to hitting a bandwidth limitation elsewhere, like the size of the non-contextual embeddings.

\begin{table}[H]
\centering
\begin{tabular}{lrrrrr}
\toprule
& \multicolumn{5}{c}{Window Count} \\
Corpus & 2 & 4 & 8 & 16 & 32 \\
\midrule
acl & \textbf{7.05} & 7.05 & 7.07 & 7.09 & 7.18 \\
greek & 6.71 & \textbf{6.66} & 6.71 & 6.70 & 6.75 \\
latin & \textbf{7.11} & 7.13 & 7.13 & 7.22 & 7.24 \\
scifi & 7.12 & \textbf{7.12} & 7.13 & 7.17 & 7.23 \\
un & \textbf{7.02} & 7.05 & 7.20 & 7.20 & 7.40 \\
\bottomrule
\end{tabular}
\caption{Average test word NLL across a range of window counts.}
\label{table:window_count}
\end{table}

Evaluating a range of window counts (Table \ref{table:window_count}) finds that performance is consistently best at small numbers ( \(2\), \(4\) ), not too surprising since the higher values produce empty windows for most corpora, leading to unnecessary steps in the random walks. However, it is striking that the performance loss from more windows is quite modest: the interpolation technique seems to be sufficient for passing over a window with no observations without catastrophic wandering. We therefore believe DETM to be fairly robust to over-granularity of time slices, and pursuing this thread to the point of approaching a continuous-time variant is an exciting prospect.

\begin{table}[H]
\centering
\begin{tabular}{lrrr}
\toprule
& \multicolumn{2}{c}{Vocab Size} \\
Corpus & 5000 & 20000 & 80000 \\
\midrule
acl & 7.05 & 7.06 & 7.05 \\
greek & 6.68 & 6.67 & 6.66 \\
latin & 7.15 & 7.14 & 7.15 \\
scifi & 7.13 & 7.14 & 7.14 \\
un & \textbf{7.10} & 7.10 & 7.17 \\
\bottomrule
\end{tabular}
\caption{Average test word NLL across a range of vocabulary sizes.}
\label{table:vocabulary_size}
\end{table}

As hoped due to its use of embeddings, Table \ref{table:vocabulary_size} shows that DETM is very stable when scaling vocabulary size. A related technical concern arises, though, with continued scaling: the operations required to construct the normalized categorical topic distributions become extremely memory-intensive, and \( 80k \) was our upper bound even on H100 GPUs. It might be useful to incorporate approaches to approximating or restructuring standard softmax \cite{grave:2016,ruiz:2018,liang:2020}. The ability to further scale the vocabulary becomes more critical in scenarios where diachronic change, orthographic variation, pre-standardized spelling, etc multiply the word-space in meaningful ways.

\begin{table}[H]
\centering
\begin{tabular}{lrrrrr}
\toprule
& \multicolumn{5}{c}{Delta Ratio} \\
Corpus & $\frac{1}{9}$ & $\frac{1}{3}$ & 1 & 3 & 9 \\
\midrule
acl & \textbf{7.03} & 7.04 & 7.07 & 7.11 & 7.23 \\
greek & 6.70 & 6.71 & \textbf{6.68} & 6.68 & 6.75 \\
latin & 7.22 & 7.17 & 7.15 & \textbf{7.13} & 7.16 \\
scifi & \textbf{7.10} & 7.12 & 7.13 & 7.15 & 7.21 \\
un & 7.07 & \textbf{7.05} & 7.09 & 7.23 & 7.32 \\
\bottomrule
\end{tabular}
\caption{Average test word NLL by ratio of mixture-walk to topic-walk delta values.}
\label{table:delta_ratio}
\end{table}

Comparison of different ratios between the deltas for the mixture and topic random walks (Table \ref{table:delta_ratio}) is not easily interpreted: the highest ratio is consistently the worst, but the best performance is scattered across the remaining four. It may simply be that the inference networks in question can readily adapt to the deltas.

Finally, we make the qualitative observation (similar to e.g. \cite{antoniak:2022}) that the output from these models should be considered intentionally: particularly with larger vocabularies, topics naturally may be dominated by inflections, the most-common terms, and so forth. By carefully defining the statistical properties of an "interesting" word/document/topic/window/etc prior to experimentation, interpretable patterns can be surfaced with a minimum of ad hoc customization (or, for humanists, premature rejection of surface-level triviality). For instance, applying formulae from \cite{sirin:2024} we find that all models trained on the scifi data capture the heterogeneity of adverb topic-membership over time, perhaps as scientific style was incorporated into popular writing, and the semantic stability of specialized and newly-coined terminology. In both cases, the extrema never appear in the top-N lists of any topics: future work will expand the inventory of general, interpretable orderings of the types of entities represented in model output, and relationships between them (e.g. authors according to idiosyncratic vocabulary w.r.t. a given topic centroid, or word-pairs closest in embedding space but furthest in temporal co-occurrence).

\section{Summary and next steps}

The performance effects mapped out in this paper are intended as an initial guide for using and continuing to develop this useful class of models. To facilitate both, we have developed a new pip-installable library to keep our implementations of variants on DETM separate from the experimental code linked above, so they continue to evolve as a pip-installable library \footnote{\url{https://github.com/ANONYMIZED}}.

In addition to the ideas noted in-line in Section \ref{sec:results}, we are pursuing several major lines of inquiry within the linked code base. The most ambitious is to design a variation on DETM that does away with discrete temporal windows in favor of a continuous random walk, e.g. by incorporating work on neural architectures for temporal event streams \cite{mei:2017}. A starting point for this is to simply allow for non-uniformly-spaced windows, to replace our interpolation workaround. A second line of inquiry is to expand the suite of robust, interpretable measurements of the entities immanent in a trained model. These can be motivated linguistically or historically, or more mechanically, e.g. to modulate observed topic collapse or instability with information from the topic-mixture distributions that may reflect whether there is a genuine issue, or if the topic is unnecessary capacity capturing the residuals of the robust parts of the model.

\section{Limitations}
NLL's straightforward measurement of predictive performance is not considered a direct proxy for common approaches to \emph{human topic interpretability}. We also note that our experiments only explore a subset of hyper-parameters and possible values. While our corpora are quite heterogeneous in terms of time, language, and style for the sake of general, coarse-grained insight, more nuanced patterns (e.g. related to linguistic typology or absolute temporal scope) may require additional materials that interpolate along these axes.
\newpage

\bibliography{detm_survey}

\appendix

\section{Experimental settings}
\label{appendix:settings}

These are the default values used for the hyper-parameters \emph{not} under consideration for the current suite of experiments:

\begin{table}[H]
\centering
\begin{tabular}{ll}
\hline
Name & Value \\
\hline
Topic count & 50 \\
Window count & 8 \\
Vocabulary size & 10000 \\
Max subdoc tokens & 100 \\
Optimizer & RAdam \\
Batch size & 512 \\
Learning rate & 0.005 \\
Epochs & 50 \\
Loss reweighting & False \\
Summary statistics recomputation & False \\
Max word sub-occurrence & 0.5 \\
\hline
\end{tabular}
\caption{Default training hyper-parameters}
\label{table:defaults}
\end{table}

\section{Coherence measurements}
\label{appendix:coherence}

Below we include tables corresponding to those reporting NLL in the main body of the paper, but instead reporting the average NPMI across the topics from all windows learned by the given model. These are computed with respect to the test corpus, using 10-word contexts and the top 20 words from each topic distribution. Ranking the models in each row according to average NPMI, and comparing to the inverse ranking by NLL, there is a weak average positive correlation of \( 0.23 \). Note the many ways this differs from the typical application of coherence metrics to rank \emph{topics} within a given model, typically inferred from preprocessed data with lemmatized and filtered vocabulary, and compared to \emph{human} annotation.

\begin{table}[H]
\centering
\begin{tabular}{lrr}
\toprule
 & \multicolumn{2}{c}{Reweight} \\
Corpus & False & True \\
\midrule
acl & -0.001 & 0.019 \\
greek & -0.000 & -0.023 \\
latin & -0.024 & -0.042 \\
scifi & 0.056 & 0.045 \\
un & -0.090 & 0.001 \\
\bottomrule
\end{tabular}
\end{table}

\begin{table}[H]
\centering
\begin{tabular}{lrr}
\toprule
 & \multicolumn{2}{c}{Recompute} \\
Corpus & False & True \\
\midrule
acl & 0.018 & 0.018 \\
greek & 0.000 & 0.002 \\
latin & -0.030 & -0.032 \\
scifi & 0.062 & 0.055 \\
un & -0.108 & -0.103 \\
\bottomrule
\end{tabular}
\end{table}

\begin{table}[H]
\centering
\begin{tabular}{lrrrrr}
\toprule
& \multicolumn{5}{c}{Delta Ratio} \\
Corpus & $\frac{1}{9}$ & $\frac{1}{3}$ & 1 & 3 & 9 \\
\midrule
acl & 0.028 & 0.023 & 0.000 & -0.046 & -0.088 \\
greek & -0.001 & 0.003 & -0.001 & -0.011 & -0.018 \\
latin & -0.016 & -0.014 & -0.023 & -0.042 & -0.098 \\
scifi & 0.065 & 0.063 & 0.054 & 0.056 & -0.007 \\
un & -0.064 & -0.071 & -0.095 & -0.167 & -0.192 \\
\bottomrule
\end{tabular}
\end{table}

\begin{table}[H]
\centering
\begin{tabular}{lrrrrrrr}
\toprule
& \multicolumn{7}{c}{Topic Count} \\
Corpus & 2 & 5 & 10 & 20 & 40 & 80 & 160 \\
\midrule
acl & 0.049 & 0.028 & 0.053 & 0.047 & 0.017 & -0.015 & -0.096 \\
greek & -0.013 & 0.007 & -0.010 & 0.003 & 0.002 & -0.018 & -0.067 \\
latin & 0.023 & 0.014 & -0.003 & -0.009 & -0.018 & -0.045 & -0.150 \\
scifi & 0.050 & 0.044 & 0.072 & 0.079 & 0.069 & 0.003 & -0.042 \\
un & -0.001 & 0.006 & -0.008 & -0.023 & -0.091 & -0.166 & -0.258 \\
\bottomrule
\end{tabular}
\end{table}

\begin{table}[H]
\centering
\begin{tabular}{lrrrrr}
\toprule
& \multicolumn{5}{c}{Window Count} \\
Corpus & 2 & 4 & 8 & 16 & 32 \\
\midrule
acl & 0.037 & 0.020 & -0.001 & -0.009 & -0.062 \\
greek & 0.005 & 0.009 & -0.004 & -0.009 & -0.027 \\
latin & -0.005 & -0.006 & -0.032 & -0.045 & -0.082 \\
scifi & 0.062 & 0.062 & 0.061 & 0.021 & -0.032 \\
un & -0.023 & -0.078 & -0.118 & -0.145 & -0.253 \\
\bottomrule
\end{tabular}
\end{table}

\begin{table}[H]
\centering
\begin{tabular}{lrrr}
\toprule
& \multicolumn{3}{c}{Vocab Size} \\
Corpus & 5000 & 20000 & 80000 \\
\midrule
acl & 0.012 & 0.022 & 0.018 \\
greek & -0.001 & 0.004 & 0.000 \\
latin & -0.030 & -0.022 & -0.025 \\
scifi & 0.056 & 0.058 & 0.059 \\
un & -0.086 & -0.101 & -0.116 \\
\bottomrule
\end{tabular}
\end{table}

\end{document}